\documentclass[letterpaper, 10 pt, conference]{ieeeconf}  

\IEEEoverridecommandlockouts                              

\usepackage{graphics} 
\usepackage{epsfig} 
\usepackage{mathptmx} 
\usepackage{times} 
\usepackage{amsmath} 
\usepackage{amssymb}  
\usepackage{subfigure}
\usepackage{hyperref}
\usepackage{dblfloatfix}
\usepackage{comment} 
\usepackage{subfigure}
\usepackage{cite}

\title{\LARGE \bf
WareVR: Virtual Reality Interface for Supervision of Autonomous Robotic System Aimed at Warehouse Stocktaking
}

\author{Ivan Kalinov, Daria Trinitatova and Dzmitry Tsetserukou

\thanks{All authors are with Skolkovo Institute of Science and Technology, Moscow 143026, Russia 
         {\tt ivan.kalinov@skolkovotech.ru, \{daria.trinitatova, d.tsetserukou\}@skoltech.ru}}
}

\begin{document}

\maketitle
\thispagestyle{empty}
\pagestyle{empty}

\begin{abstract}

WareVR is a novel human-robot interface based on a virtual reality (VR) application to interact with a heterogeneous robotic system for automated inventory management. We have created an interface to supervise an autonomous robot remotely from a secluded workstation in a warehouse that could benefit during the current pandemic COVID-19 since the stocktaking is a necessary and regular process in warehouses, which involves a group of people. The proposed interface allows regular warehouse workers without experience in robotics to control the heterogeneous robotic system consisting of an unmanned ground vehicle (UGV) and unmanned aerial vehicle (UAV). WareVR provides visualization of the robotic system in a digital twin of the warehouse, which is accompanied by a real-time video stream from the real environment through an on-board UAV camera. Using the WareVR interface, the operator can conduct different levels of stocktaking, monitor the inventory process remotely, and teleoperate the drone for a more detailed inspection. Besides, the developed interface includes remote control of the UAV for intuitive and straightforward human interaction with the autonomous robot for stocktaking. The effectiveness of the VR-based interface was evaluated through the user study in a ``visual inspection'' scenario.

\end{abstract}

\section{Introduction}

\subsection{Motivation}
Nowadays, we are witnessing revolutionary changes in UAV technology resulting in redesigning the business models and creating a new operating environment in a variety of industries, from entertainment and meditation \cite{la2019drones} to assistance in territory monitoring \cite{yatskin2017principles}. At an early date, customers from a wide range of industries will witness the first impact of UAVs in various areas, from delivery services to power line inspection. UAV-based solutions are most relevant for industries that require both mobility and high-quality information. The integration of such systems into the daily workflow can provide significant advantages when implementing large capital construction projects, infrastructure management, agriculture, and 3D surface deformation \cite{braley2018griddrones}.  

Conducting a stocktaking of industrial warehouses in a fully autonomous mode is a highly promising and, at the same time, challenging task. As a rule, such a warehouse is equipped with a large number of racks with a height of 10-12 m. Such a height makes it impossible to use a robotic arm for this task. Each pallet and its place number is marked with a unique identifier, for example, a barcode, quick response (QR) code, or a  Radio Frequency IDentification (RFID), while the barcode is the most common type of identification. At the moment, most of the stocktaking of such warehouses is carried out manually \cite{wawrla2019applications}. The pallets are removed from the shelves, scanned, and then put back during the inventory process. This manual operation takes a tremendous amount of time and leads to warehouse downtime of up to 4 days for an average warehouse of 10,000 m$^2$ \cite{fernandez2019towards}. An autonomous UAV with an indoor localization system could be an ideal solution to this problem. However, the main problem is the precise positioning of the UAV within the narrow aisles between the racks. To overcome this problem, we developed a heterogeneous robotic system consisting of UAV and UGV. The proposed approach makes it possible to install all the necessary localization and navigation equipment on two robots, as opposed to motion capture systems, which require additional active infrastructure, which is impossible for warehouses. In addition, the UAV flight time is limited to 25 minutes, during which, on average, it is possible to carry out an inventory of no more than 200 m$^2$ of the warehouse. Therefore, we decided to use our mobile platform for recharging the UAV through contact pads.

\begin{figure}[!t]
\centering
\includegraphics[width=0.98\linewidth]{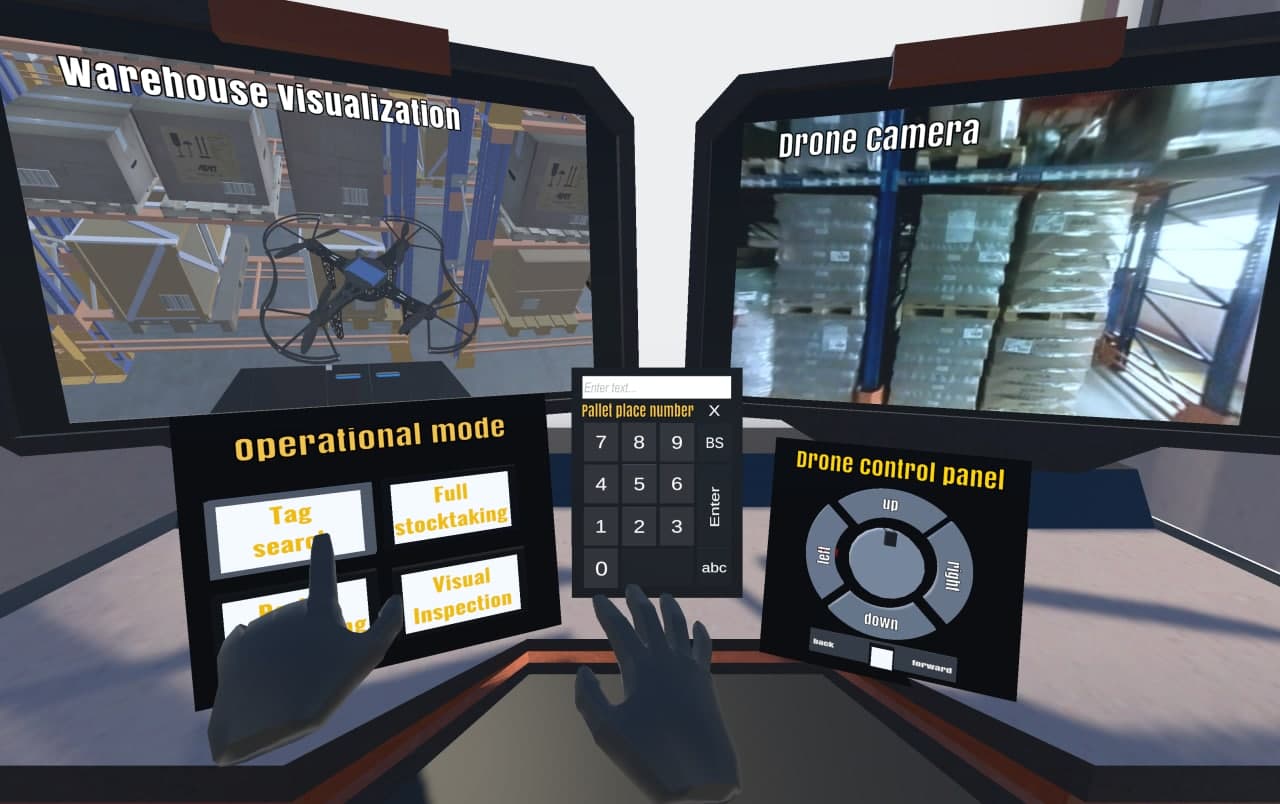}
\caption{WaveVR interface.}
\label{fig:main}
\vspace{-1em}
\end{figure}

We have developed a heterogeneous robotic system that can localize and navigate in the indoor warehouse environment autonomously \cite{kalinov2019high, kalinov2020warevision, kalinov2021impedance}. The system copes with its task and makes an inventory with photos, uploading all the information into a database and building a product location map.
In some cases, the data from the autonomous inventory (recognition result, coordinates, photo) is insufficient, and a more detailed inspection of a particular pallet is needed. This type of inspection is not available offline, but sometimes it is required promptly and remotely. Therefore, it is essential to have an intuitive human-computer interface to interact with the autonomous system with clear feedback, so that even warehouse workers inexperienced in robot control can easily handle it. Such an interface should include two operation modes for the system: automatic and manual. The use of virtual reality can provide a comfortable perception of the environment around the robot and increase the level of involvement and realism during teleoperation compared to the regular screen interface.

The proposed WareVR interface allows regular warehouse workers without experience in robotics to control the heterogeneous robotic system in the VR application, which provides visualization of the heterogeneous robotic system in the digital twin of the warehouse. Besides, our interface allows stream video from the UAV camera into the virtual environment for additional visual feedback from the real environment. The user can operate the robotic system with velocity control using the hand-held VR controller.

\subsection{Related works}

In the evolution of the Human-Computer Interaction (HCI) discipline, recently, special attention is paid to human-drone interaction (HDI). Drones are becoming more and more diffused, being used with different purposes in various control modes, e.g., automatic, manual, shared-autonomy, when the user can execute some operations while the autonomous system works. Silvia Mirri et al. \cite{mirri2019human} presented a thorough overview of the latest papers in HDI. One of the main problems with almost all control interfaces is the necessary experience in drone control for their use in challenging tasks, e.g., indoor operation \cite{erat2018drone}, risky operation \cite{aleotti2017detection}, inspection \cite{irizarry2012usability}, etc. 
One of the most common approaches to remote control of drones is the use of computer vision methods. In several works \cite{naseer2013followme, mohaimenianpour2018hands}, the operator implemented UAV control using an on-board camera (RGB camera, Kinect sensor) that recognizes face and hand gestures. However, this approach cannot be applied if the operator is out of the UAV camera field of view. 

Nowadays, one of the most prospective approaches for remote robot control is teleoperation via physical sensor-based (e.g., motion capture or electromyography) interface \cite{miehlbradt2018data}, \cite{wu2019teleoperation}. Thus, Rognon et al. \cite{rognon2018flyjacket} developed FlyJacket, a soft exoskeleton for UAV control by body motion. The exoskeleton contains a motion-tracking device to detect body movements and a VR headset to provide visual feedback. However, such interfaces can be bulky and complicated for deploying and using quickly and require preliminary operator training. Besides, the exoskeleton as a concept for robot control significantly limits the motion of the operator.

Virtual reality user interfaces provide the operator with more immersive interaction with robots. Thomason et al. \cite{thomason2019comparison} developed a virtual reality interface for safe drone navigation in a complex environment. The teleoperation system provides the user with environment-adaptive viewpoints in real-time to maximize visibility. In \cite{vempati2019virtual}, it was proposed a VR interface to control an autonomous UAV for spray painting on complex surfaces. Using a virtual spray gun, it allows the user to move around the target surface in a virtual environment and paint at desired locations. Patterson et al. \cite{paterson2019improving} presented an open-source platform for 3D aerial path planning in VR. The introduced VR interface has advantages in both usability and safety over manual interfaces and can significantly reduce path planning time compared to a 2D touchscreen interface. Yashin et al. \cite{aerovr2019} proposed a VR-based teleoperation system for an aerial manipulator. The developed system consists of a VR application with a digital twin of a drone and a wearable control interface represented by VR trackers worn on the operator's arm and a tracking glove with vibrotactile feedback to control the position of the robotic arm mounted on the UAV. 

Most of the VR teleoperation systems propose direct robot control via VR controllers or wearable interfaces. However, in our case, it is needed a semi-autonomous interface that allows choosing the operation mode of the autonomous robotic system and, if necessary, providing manual operation mode. These needs can be met with an intuitive graphical user interface (GUI) in VR.

\begin{figure*}[!h]
\centering
\includegraphics[width=0.95\textwidth]{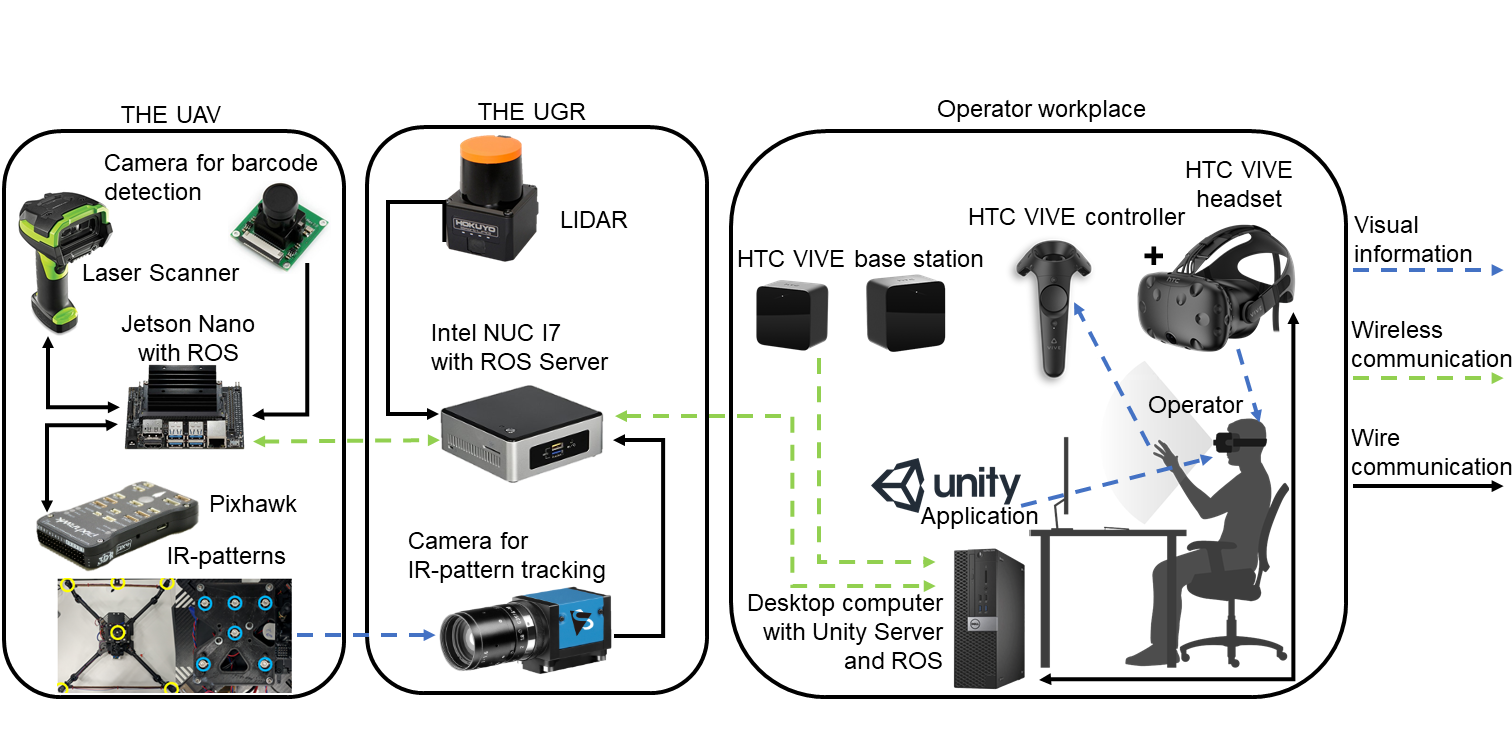}
\vspace{-1em}
\caption{The overall architecture of the developed system. The key equipment of the heterogeneous robot, server information flow using ROS, and operator layers are presented in the form of block diagrams. }~\label{fig:scheme}
\vspace{-1em}
\end{figure*}
\section{System Overview}

Fig. \ref{fig:scheme} shows the overall architecture of the developed system, which includes the key equipment of the heterogeneous robot, Robot Operating System (ROS) framework, Unity, and the operator workplace. We used the HTC VIVE headset to represent a virtual environment from the Unity application and the HTC VIVE controller to track user hand position to control the system. The desktop computer with installed ROS and Unity is located in the operator's workspace, the mobile platform is controlled by an on-board computer based on the Intel NUC with Core i7 processor, the on-board computer of the UAV is Nvidia Jetson Nano, which directly interacts with the Pixhawk flight controller. All three computers operate in multi-master mode in ROS to ensure reliable information transfer within the one local Wi-Fi network in the warehouse. The connection between ROS and Unity application is based on the  ROS-Unity Communication Package (ROS$\#$). 

\subsection{Operation principles of heterogeneous robotic system}

The main objective of the UGV is to determine its coordinates relative to the surrounding objects in the warehouse and accurately calculate the position of each pallet. To do this, we use modern approaches for robot localization and navigation. The robot uses a simultaneous localization and mapping (SLAM) algorithm based on graph-based approaches \cite{Hess2016} to determine its location in space. For map building of the surveyed area, the UGV employs a Light Detection and Ranging device (LIDAR). 


To detect and recognize UAV position relative to the UGV, we installed the camera on the ground robot and two concentric patterns of active infrared (IR) markers on the UAV.  
In our setup, we employ a global shutter ImagingSource camera with a resolution of 2448x2048 and a lens with 137.9$^\circ$ field of view (FoV). Besides, we use an IR-passing 950 nm filter to obscure all the light (e.g., ceiling backlight) except for IR markers. All the key hardware components and communication between them are represented in Fig. \ref{fig:scheme}. The detailed explanation of the system setup is described in our previous work \cite{kalinov2019high}. 


\subsection{Barcode mapping}
\label{mapping}
For acquiring augmented UAV localization, we obtain the exact coordinates of the barcode relatively to the UAV and verify them by a laser scanner, which makes it possible to use it as an additional source of information. At the first stage, we get data from the convolutional neural network (CNN) U-Net on the possible position of barcodes in the image frame \cite{kalinov2020warevision}. Based on the position of barcodes, we make a preliminary barcode map and generate our preliminary global flight path. Then the UAV starts scanning along this path. The area where the barcode was detected and verified by the laser scanner is assigned a ``verified barcode'' status and gets into the barcode database, after which the drone flies to the next waypoint. If the laser scanner cannot read the barcode, in this case, our algorithm removes this waypoint from the path and assigns the next waypoint to the drone \cite{kalinov2020warevision}. 

In addition, during the flight, we obtain the linear dimensions of the boxes with an accuracy of 3 cm using the distance to the boxes and detecting their contours by the standard OpenCV functions. This accuracy allows to classify boxes by typical sizes and use them for a digital twin recreation \autoref{twin}. The recognized barcodes are linked to the boxes according to their position.

\begin{figure}[h]
\centering
\includegraphics[width=0.9\linewidth]{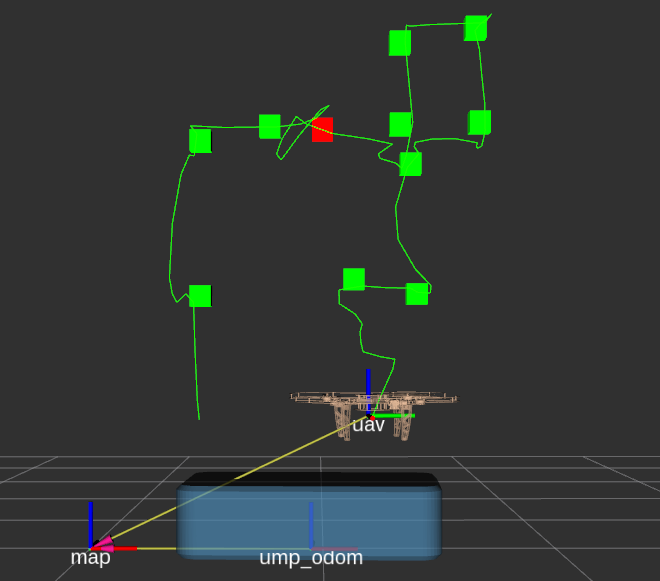}
\caption{The accurate map of barcodes in RVIZ environment. The green boxes represent verified barcodes. The red box represents false-positive detected barcode with CNN.}~\label{fig:marginfig}
\end{figure}

The result of this work is an accurate map of recognized barcodes and goods in a warehouse (Fig. \ref{fig:marginfig}).

\subsection{VR system}
We have developed a VR interface based on the Unity game engine, which allows monitoring and managing the robot inventory process directly in real time. The VR setup includes HTC Vive Pro base stations, head-mounted display (HMD), and HTC VIVE controller for tracking the position of the user’s hand in VR. The 3D graphical user interface (GUI) developed by us includes a panel for control of a full inventory process, as well as a manual operation of the UAV, one screen for providing visual feedback from the drone camera, and another for representing a digital twin of the warehouse (Fig. \ref{fig:main}). The control board consists of three panels: a panel for choosing the operational mode, a panel for the input of pallet place number, and a panel for the manual UAV control. 

\section{VR Interactive Interface}

\subsection{A digital twin of the warehouse}
\label{twin}
\begin{figure}[h]
\centering
    \includegraphics[width=0.95\linewidth]{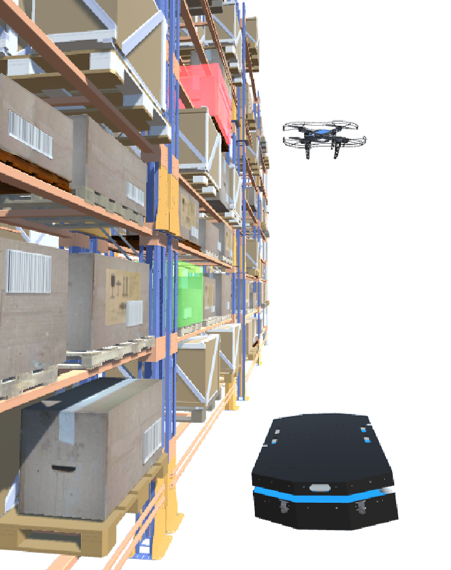}
    \caption{Visualization of the UAV flight above the UGV near one rack in the warehouse digital twin. In our user study, we used pallet highlighting for scanning as follows: the green box indicates a scanned pallet, the red box indicates a pallet that requires scanning. }~\label{fig:wareVR}
\end{figure}

The left screen on the GUI is used to visualize the target warehouse in the VR environment (Fig. \ref{fig:main}). 
Initially, the digital twin of the warehouse is generated based on the given input parameters. This list includes a 2D map of the warehouse walls, the ceiling height, the number of racks and their initial positions, the distance between racks, the number of tiers and sections in each rack, the size of the rack cells (pallet places), and a list of typical boxes with dimensions. Then we place the heterogeneous robot in our model of the digital twin. Filling the shelves with pallets and boxes occurs based on the created map of detected and verified barcodes during the stocktaking (\autoref{mapping}).
When the UAV detects a barcode on the real goods, a virtual pallet is filled with a box in the digital warehouse only after barcode verification. In addition, at the time of verification, the UAV takes a photo of the pallet and links it in the database with the scanned barcode. Thus, we reconstruct a simplified model of the inspected warehouse (Fig. \ref{fig:wareVR}).

\subsection{Safe operating modes}

Although one of the advantages of the proposed system is the possibility of remote use, which eliminates the risk of the drone operation near the operator, nevertheless, other employees may work in the warehouse during this time. Therefore, from the point of view of human-robot interaction, it is essential to describe the modes of safe operation. We utilize all the data used to create the digital twin (2D map of the warehouse walls, the ceiling height, the initial positions of the racks, etc.) in the robot control system. Thus, we generate a 3D map of the space with the zones where the drone can fly. This is done to significantly reduce the risk of the drone collision with the racks. It is also important to note that the drone is always located in the cylindrical area above the mobile robot. In case of any connection loss or an abnormal situation, the drone will provide a soft landing on the mobile robot \cite{kalinov2021impedance}.

\subsection{Operational modes}
The control panel includes four operational modes of stocktaking for the operator: 

\begin{itemize}
\item \textbf{Full stocktaking}. During this mode, the system works fully automated and conducts stocktaking of the whole warehouse. In this mode, the operator supervises the system and visually checks the conditions of the pallets (damaged, opened, etc.) using a video stream from the UAV camera. At any time, the operator could pause the stocktaking to zoom in the image from the UAV and check the pallet condition and then give a command to continue stocktaking in automatic mode or abort stocktaking.
\item  \textbf{Partial stocktaking}. During this mode, the system works fully automated and conducts stocktaking of a selected part of the warehouse. The operator supervises the system and could check the pallet as in ``Full stocktaking mode''.
\item  \textbf{Tag search}. The system works fully automated during this mode and searches a defined tag using the information from the previous stocktaking as the initial parameters. If the pallet was not found in its past location, the system automatically gradually increases the search area within the same alley. If no pallet is found in this alley upon completion of the search, the system offers to select another alley for searching or switch to ``Visual inspection mode''.
\item \textbf{Visual inspection}. During this mode, the system is controlled manually by the operator. The first step for the operator is to enter the target location for the inspection. The robotic system autonomously arrives at the desired row, and the drone takes off at the desired height. Then the operator can guide the drone manually closer to the rack at different angles and adjust the view.
\end{itemize}

\subsection{UAV teleoperation}
\label{teleop}
A drone control panel allows the user to operate the UAV in manual mode. It comprises four buttons for the translational positioning of a drone along $X$ (left, right) and $Z$ (up, down) axes. Besides, there is an opportunity to zoom in and out the distance between the UAV and the rack with a slider (movement along the $Y$-axis). All buttons are selected with the HTC VIVE controller. For velocity control of the UAV without panel buttons, we track the position of the HTC VIVE controller by HTC VIVE base stations. To control the 4 DoF of the UAV, we use four inputs obtained from the HTC VIVE controller ($x_c$, $y_c$, $z_c$, yaw) to calculate the velocities of the UAV along the three axes as well as the target UAV yaw angle. The controlled UAV responds to changes in the position of the controller in the following way. If the HTC VIVE controller is moved horizontally (forward, backward, left, or right), the drone is commanded to change its velocity in the horizontal plane proportionally to the controller displacement. Controlling the movement along the $Z$-axis works the same way. Besides, We used the trackpad button for drone rotation in the yaw direction.

\subsection{Visual Feedback}
The right screen on the GUI in the VR application is used for providing visual feedback during the work of the robotic system from the inspection place. A video from the UAV camera is streaming in real-time to the predefined IP address in the local network during the inventory process. Unity application displays this video stream from the URL source to the interface screen.

\section{User Study}

We conducted an experiment with UAV control in a simulated environment using VR and desktop applications to evaluate the convenience of control modes and identify possible emergencies with the real system.

The purpose of the user study was to evaluate the effectiveness of the proposed approach of the manual UAV control through the developed VR interface in comparison with one of the most popular methods of UAV control using First Person View (FPV) glasses and a remote controller (RC) \cite{grijalva2019landmark}. Since usually only the image frame from the drone camera is transmitted in FPV glasses, in this mode, it is impossible to use the advantages of the VR interface, e.g., 360$^\circ$ view of the environment through the head rotation. Therefore, we simulated this control mode by streaming the image from the drone camera from the digital twin to the desktop monitor. Thereby, we would like to compare desktop and VR-based applications to prove the effectiveness of using the VR mode in our work.  

\subsection{Experimental Description}

\begin{figure}[h]
\begin{center}
\subfigure[Desktop mode.]{
\includegraphics[width=0.9\linewidth]{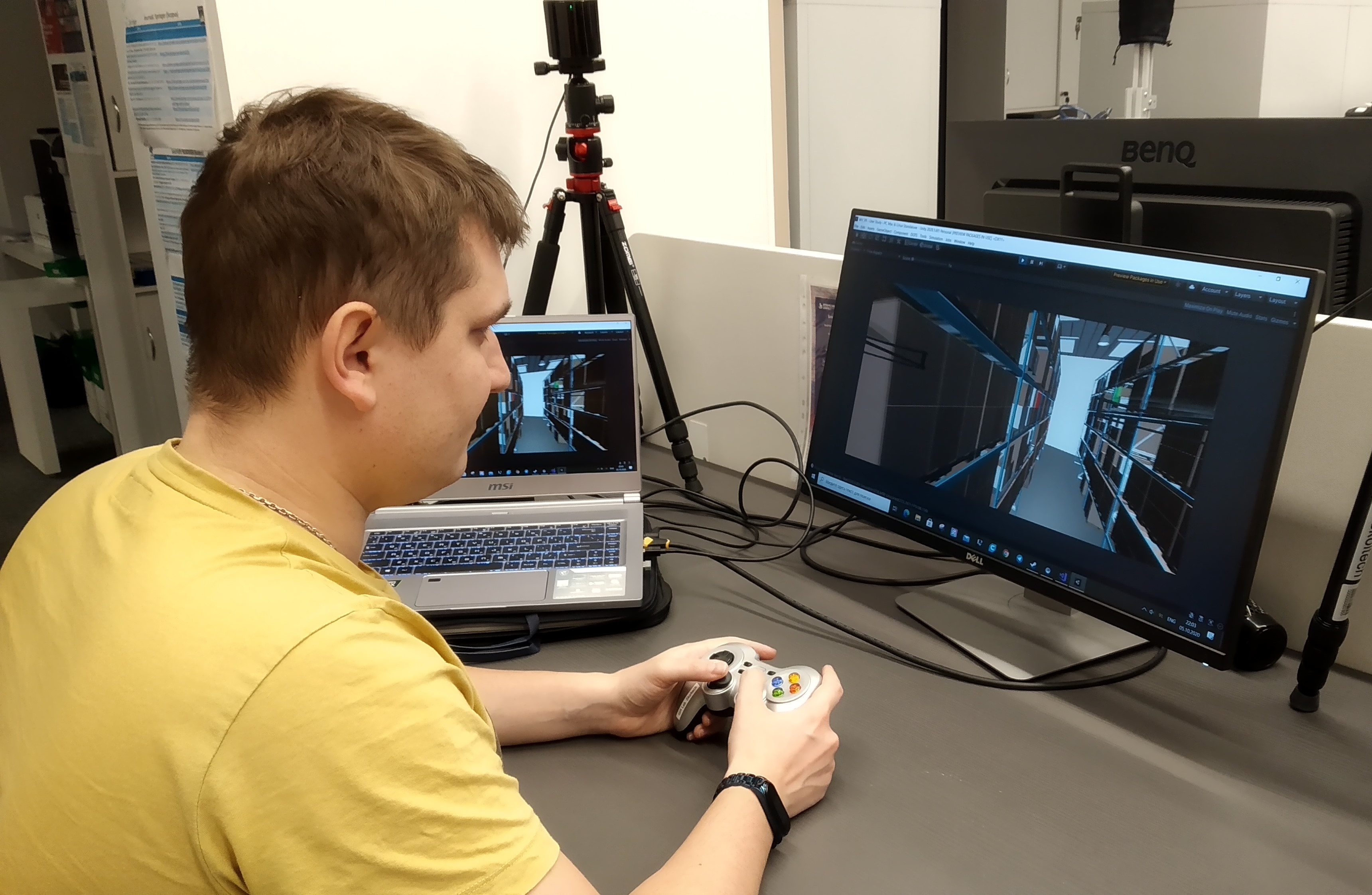}}
\subfigure[VR mode.]{
\includegraphics[width=0.9\linewidth]{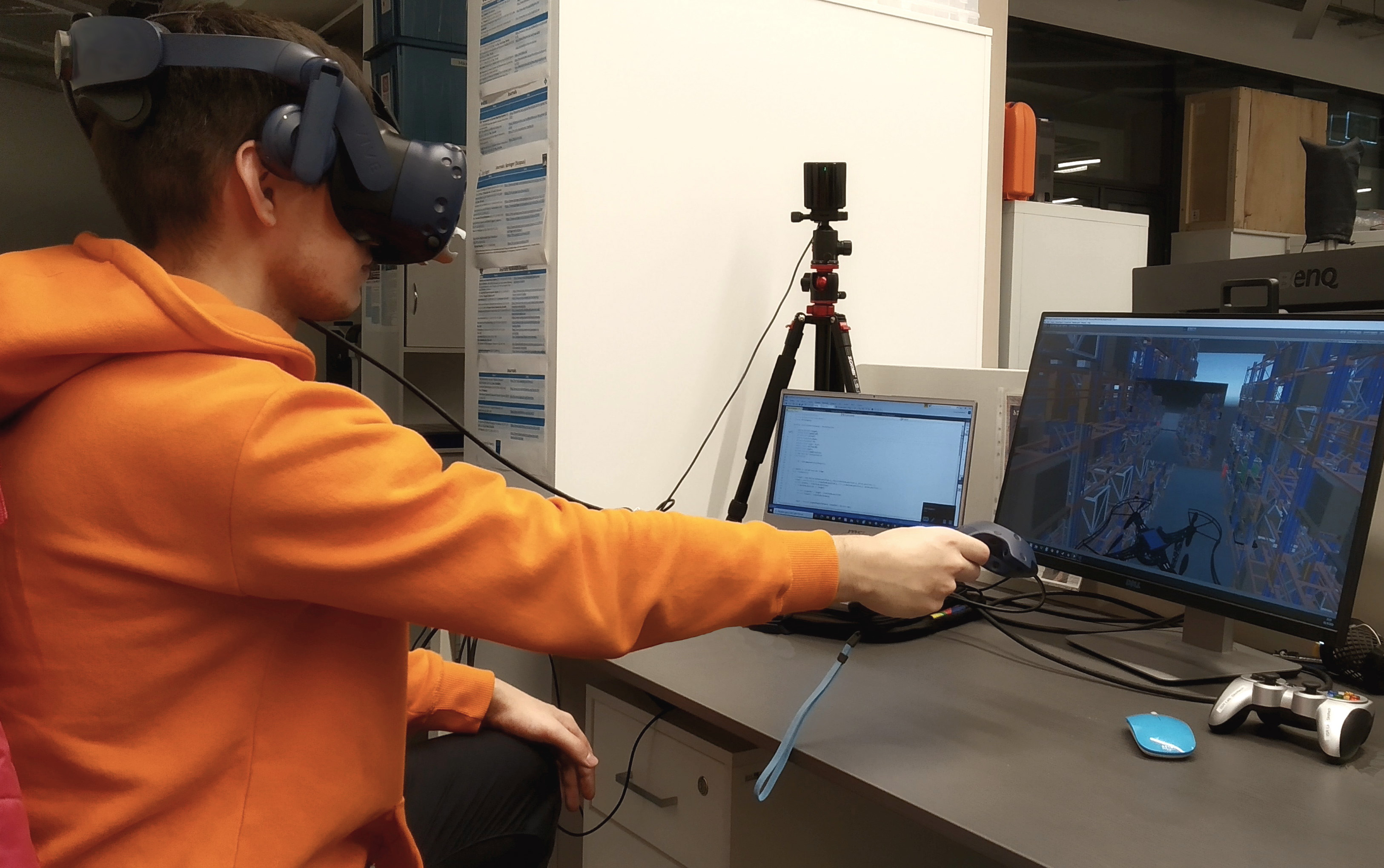}}
\caption{Overview of the experiment process.}\label{fig:participants}
\end{center}
\vspace{-2em}
\end{figure}

For the user study, the following scenario was developed. The system was launched in the ``visual inspection'' mode. The participant had to find 5 pallets and fly up to them, thereby making an inventory. All pallets that the user had to scan during the experiment, we highlighted in a red box. After scanning, the highlight color changed to green (Fig. \ref{fig:wareVR}). To simulate a flight in FPV mode, we broadcast the video stream from the drone camera from the digital environment to the desktop display (Fig. \ref{fig:participants} (a)). Flight control was carried out using two thumbsticks on the Logitech Gamepad F710 (the left thumbstick for the torque control and yaw rotation, the right one for movement in the horizontal direction: forward, backward, left, or right). This setup was chosen to simulate the most common drone control using an RC. No other camera views were available in this mode, the only video from the drone camera, as in normal control through FPV glasses. In the VR scenario with the developed interface, the UAV was controlled using the HTC VIVE controller (\autoref{teleop}). The trigger button was used to move the UAV to the holding position mode allowing the user to rotate the UAV in the yaw direction only. The subjects controlled the quadcopter from the third-person view and could rotate the head to look around the space (Fig. \ref{fig:participants} (b)). Besides, an additional window with the drone camera view was available for the user.

\begin{figure*}[!h]
\centering
\includegraphics[width=0.99\textwidth]{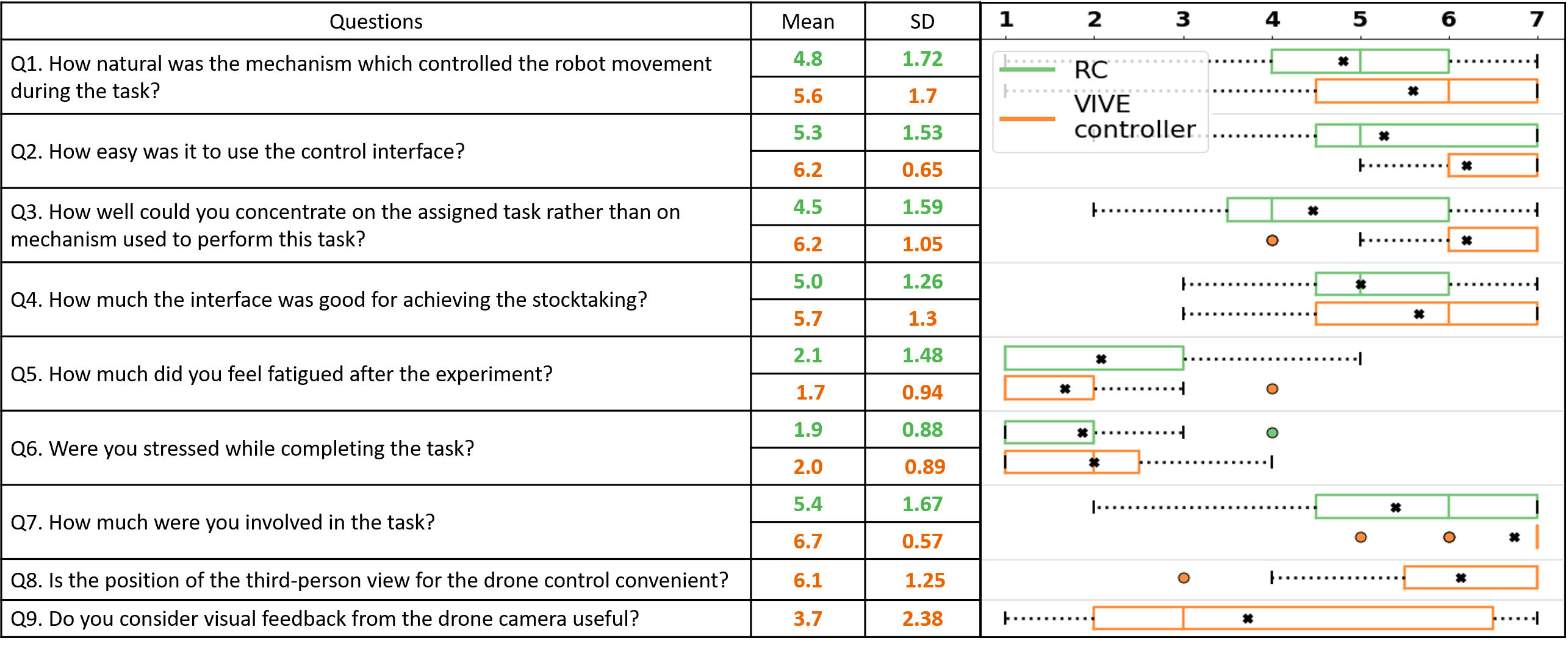}
\caption{Evaluation of the participant’s experience for both control methods in the form of a 7-point Likert scale (1 = completely disagree, 7 = completely agree). Means and standard deviations (SD) are presented. Crosses mark mean values.} 
\label{fig:Likert}
\end{figure*}

Before the experiment, each subject had short training to get acquainted with the control procedure.
After completing the tasks using both modes, we asked participants to respond to a 9-question survey using a seven-point Likert scale. The survey results are presented in Fig. \ref{fig:Likert}.

\subsection{Participants}
In total, 15 subjects took part in the experiment (3 women and 12 men). Participants were students with a background in mechanical engineering, computer science, and robotics. The average participant age was 24.4 ($SD = 2.4$), with a range of 21–31. Our sample of participants included both novice users and experienced users at drone piloting. In total, 3 participants had never interacted with drones before, 7 participants piloted drones several times, and 5 reported regular experience with aerial robots. As for VR experience, 11 participants used VR only a few times, and 4 people answered that they used VR devices regularly.

\subsection{Results and discussion}
Overall, the subjects assessed the ease of the UAV control process (Q1, Q2, Q3). The participants noted that the control mechanism was more natural (Q1) and easier (Q2) in the case of using a VR mode (Q1: $\mu=5.6,\ SD=1.7$ for VR and $\mu=4.8,\  SD=1.72$ for the RC control mode; Q2: $\mu=6.2,\ SD=0.65$ and $\mu=5.3,\ SD=1.53$ for VR and RC control, respectively). Using the one-way ANOVA, with a chosen significance level of $p<0.05$, we found a statistically significant difference in the ease of control between the two interfaces $(F(1,28)=4.43,\ p=0.04<0.05)$. According to ANOVA results, the type of control interface affects the involvement in the task (Q7) ($F(1,28)=8.02,\ p=8.5\cdot10^{-3}<0.05$) and the ability to concentrate (Q3) on the task ($F(1,28)=11.66,\ p=2\cdot10^{-3}<0.05$). In addition, for the VR mode, the subjects noted the convenience of the third-person view during the robot control (Q8: $\mu = 6.1$, $SD=1.25$).

\begin{figure}[!h]
\centering
\includegraphics[width=1\linewidth]{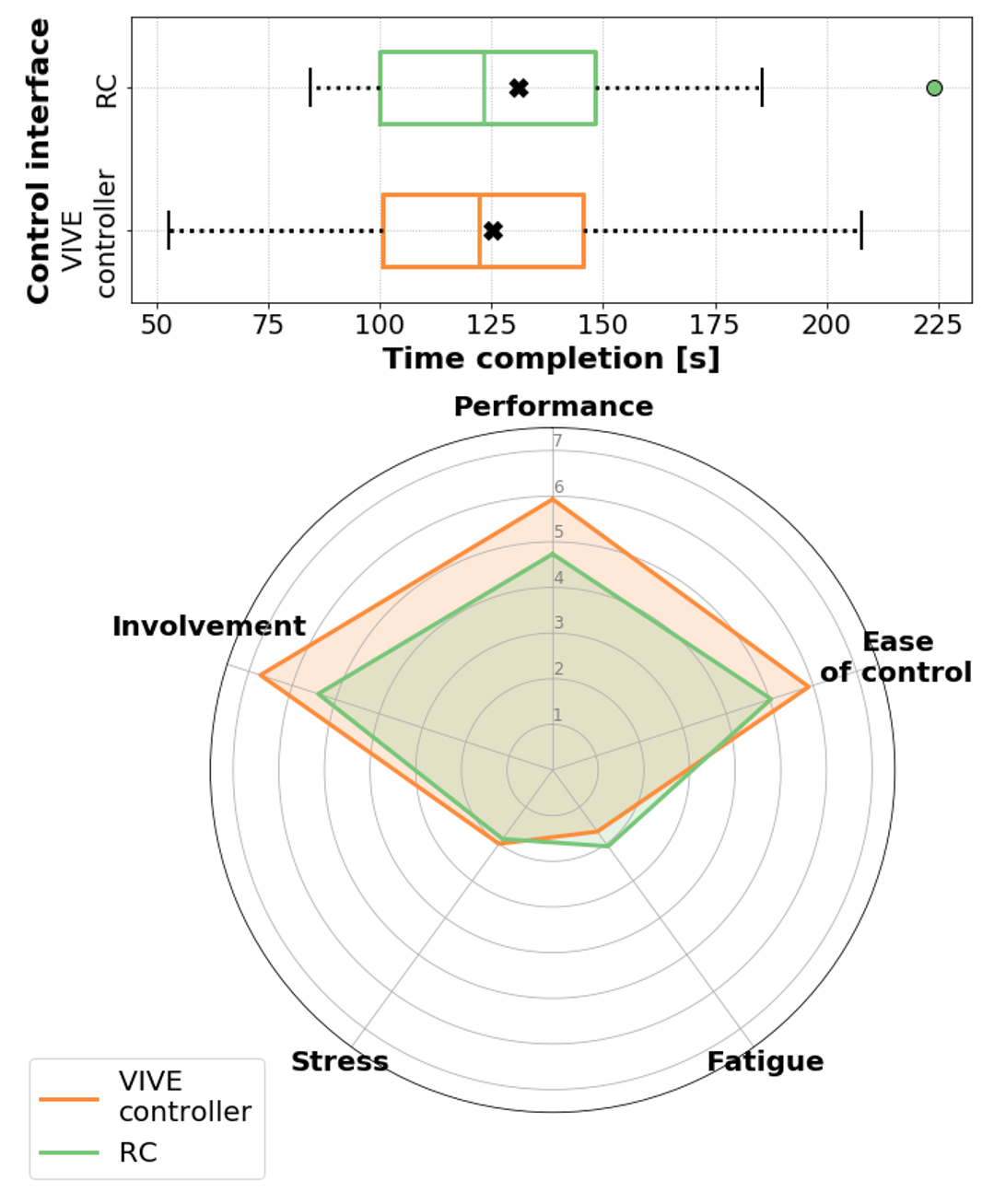}
\caption{Experimental results averaged across all participants. Top: completion time; crosses mark mean values. Bottom: radar chart. } 
\label{fig:US}
\vspace{-1em}
\end{figure}

Fig. \ref{fig:US} shows the results for task completion time and evaluation of control modes in the form of a radar chart across all participants. The difference in completion times is insignificant ($F(1,28) = 0.17,\ p = 0.68$) according to one-way ANOVA. Thus, the task execution time was comparable for both control modes (the average completion time is 130.8 $s$ with $SD=36.98$ for the desktop mode and 125.1 $s$ with $SD=37.1$ for the VR mode). The radar chart shows the user evaluation of two control modes in terms of Performance (Q3, Q4), Ease of control (Q1, Q2), Stress (Q6), Fatigue (Q5), and Involvement (Q7). According to the results of the user study, VR-based interface increases involvement and performance, whereas the stress level during task completion and fatigue of participants were almost the same for both control modes. At the same time, participants noted that it is easier to control the system through the VR application. 

\section{Conclusions and Future Work}
We have proposed a novel interactive interface based on VR application for natural and intuitive human interaction with the autonomous robotic system for stocktaking. It allows the operator to conduct different levels of stocktaking, remotely monitor the inventory process, and teleoperate the drone for a more detailed inspection. According to the user study results, the VR control interface showed a better performance than the FPV mode in terms of ease of drone control and involvement in the task. It could be concluded that WareVR suggests a new way of communication between the robotic system and operator and can potentially improve and facilitate the inventory process.\par

\begin{figure}[h]
\centering
  \includegraphics[width=0.95\linewidth]{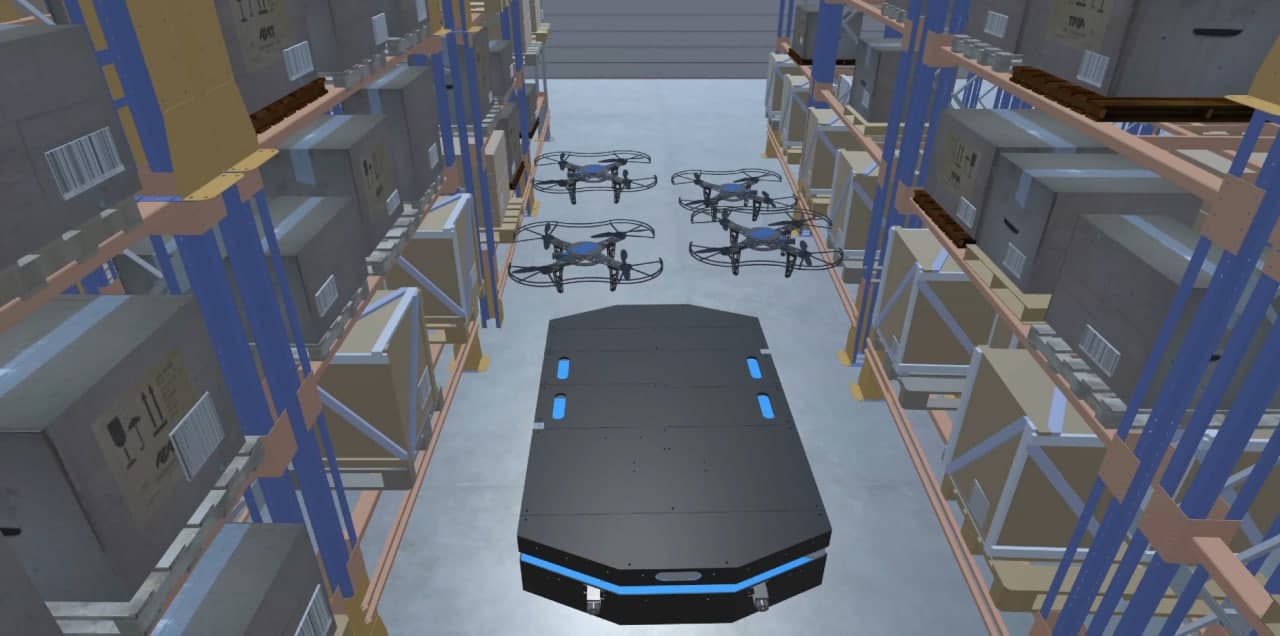}
  \caption{The heterogeneous robotic system with swarm of drones in the digital twin environment.}~\label{fig:swarm}
\vspace{-1em}
\end{figure}

In future work, we are going to:
\begin{itemize}
\item Incorporate a tracking glove with tactile feedback for the operator instead of using the HTC VIVE controller. It will facilitate the UAV control by introducing an understanding of the surrounding environment, working area, and obstacles. 
\item Add the functionality of the gesture control system to control not only a single drone but a whole swarm of drones (see Fig. \ref{fig:swarm}). Conducting experiments on gesture recognition with machine learning (ML).
\item Currently, only individual tests of real flights in the warehouse have been conducted, and tests for remote control with the developed interface in the laboratory. The next stage will be testing the whole integrated system in the warehouse and conducting experiments with warehouse employees.

\end{itemize}

\section*{Acknowledgements}

The reported study was funded by RFBR, project number 20-38-90294.


\bibliographystyle{IEEEtran}
\bibliography{ref}

\end{document}